\def\year{2019}\relax
\documentclass[sigconf]{acmart}
\usepackage{times}  
\usepackage{helvet}  
\usepackage{courier}  
\usepackage{url}  
\usepackage{graphicx}  
\usepackage[utf8]{inputenc} 
\usepackage[T1]{fontenc}    
\usepackage{hyperref}       
\usepackage{url}            
\usepackage{booktabs}       
\usepackage{amsfonts}       
\usepackage{nicefrac}       
\usepackage{microtype}      
\usepackage{subcaption}      
\usepackage{multirow,array}
\usepackage{float}
\usepackage{algorithm}
\usepackage[noend]{algpseudocode}
\usepackage{float}
\usepackage{graphicx} 
\usepackage{amsmath}
\usepackage{amssymb}

\title{Reinforcement Learning and Inverse Reinforcement Learning with System 1 and System 2}

\author{Alexander Peysakhovich}
\affiliation{%
  \institution{Facebook AI Research}
}
\email{alexpeys@fb.com}

\renewcommand{\shortauthors}{Peysakhovich}
\keywords{reinforcement learning, behavioral economics, dual-system models, inverse reinforcement learning}

\copyrightyear{2019} 
\acmYear{2019} 
\setcopyright{acmcopyright}
\acmConference[AIES '19]{2019 AAAI/ACM Conference on AI, Ethics, and Society}{January 27--28, 2019}{Honolulu, HI, USA}
\acmBooktitle{2019 AAAI/ACM Conference on AI, Ethics, and Society (AIES'19), January 27--28, 2019, Honolulu, HI, USA}
\acmPrice{15.00}
\acmDOI{10.1145/3306618.3314259}
\acmISBN{978-1-4503-6324-2/19/01} 

\begin{document}

\begin{abstract}
Inferring a person's goal from their behavior is an important problem in applications of AI (e.g. automated assistants, recommender systems). The workhorse model for this task is the rational actor model - this amounts to assuming that people have stable reward functions, discount the future exponentially, and construct optimal plans. Under the rational actor assumption techniques such as inverse reinforcement learning (IRL) can be used to infer a person's goals from their actions. A competing model is the dual-system model. Here decisions are the result of an interplay between a fast, automatic, heuristic-based system 1 and a slower, deliberate, calculating system 2. We generalize the dual system framework to the case of Markov decision problems and show how to compute optimal plans for dual-system agents. We show that dual-system agents exhibit behaviors that are incompatible with rational actor assumption. We show that naive applications of rational-actor IRL to the behavior of dual-system agents can generate wrong inference about the agents' goals and suggest interventions that actually reduce the agent's overall utility. Finally, we adapt a simple IRL algorithm to correctly infer the goals of dual system decision-makers. This allows us to make interventions that help, rather than hinder, the dual-system agent's ability to reach their true goals.
\end{abstract}

\maketitle

\section{Introduction}
Modeling human decision making and inferring a person's latent reward function from their behavior are important problems across many fields \cite{resnick1997recommender,athey2010structural,keren2014goal,hadfield2016cooperative}. Typically such inference is performed using the rational actor model. The rational actor model assumes that people have a fixed utility function (aka. reward function), discount the future exponentially, and that they are capable of planning. Assuming the rational actor model means that if we know a person's reward function we can predict their behavior they will take by using dynamic programming to find (approximately) optimal plans \cite{sutton1998introduction}. Similarly, if we have observations of a person's behavior we can invert this behavior to learn the goals they are trying to achieve \cite{ng2000algorithms,ziebart2008maximum,ramachandran2007bayesian}. 

Unfortunately, it is well known that the rational actor model fails in many important decision environments \cite{thaler2012winner}. The goal of this paper is to ask whether the tools of planning and inverse planning can be applied to the dual-system model (2S), a workhorse model of human decision-making from the behavioral and cognitive sciences \cite{kahneman2011thinking}.

The 2S views behavior as being controlled by two systems, system 1 which is automatic, fast, effortless and uses heuristics and system 2 which is slower, reflective, requires cognitive costs and deliberates. This paper follows existing work in neuroscience and assumes an interaction between systems as follows: when faced with a decision first system 1 `suggests' a course of action and system 2 uses costly cognitive control to modulate this suggestion \cite{hare2009self,shenhav2013expected}. 

An important example of violations of the rational actor model occur in situations where people must trade off rewards now for rewards later \cite{ainslie2001breakdown}. People plan to eat healthy, go to the gym, and quit smoking \textit{tomorrow} but when tomorrow comes, they reverse their plans \cite{o1999doing,thaler1981economic}. Patterns of choices where one alternative is chosen when the choice is made for the future but the other is chosen when the choice is made for immediate outcomes (e.g. committing to start eating healthy tomorrow but having a donut today) are called dynamically inconsistent. Dynamic inconsistency cannot occur if individuals are indeed maximizing a stable utility function which discounts the future exponentially \cite{o1999doing}.\footnote{Another pattern of real world behavior that points to violations of the rational actor model is the presence of commitment devices \cite{bryan2010commitment,peysakhovich2014commit}. People are willing to pay to remove choices from their choice sets (e.g. pay a personal trainer to force them out of bed at 6am for a workout). A rational agent (who can simply choose to follow through on any plans made yesterday) would never make such a decision.}  

The 2S model states that time inconsistency arises because decisions are an interplay between a system 1 that seeks immediate gratification and a system 2 that is able to consider long term impacts of decisions \cite{thaler1981economic,mcclure2004separate,fudenberg2006dual}. When decisions are only about the future, system 2 wants to commit to eating healthy and going to the gym but when the donut is in front of us system 1 makes it hard to put down.

This paper asks: what happens when decisions are not between single actions but in temporally extended plans? We will consider a computational model of the 2S planning as follows: the decision-maker faces a Markov decision problem and the resulting choices are a product of system 2 optimizing some utility function net of cognitive control costs. The cognitive control cost of a decision is proportional to how much it deviates (in terms of disutility) from an optimal decision that system 1 would like to make.\footnote{Throughout the paper language like `desires', `beliefs', and `conflict' will be used to anthropomorphize system 1 and 2. This is not meant to imply the existence of homunculi literally fighting within an individual's head, rather the language is used to convey intuitions behind concepts.} System 1 and system 2 have different utility functions, discount the future differently, or both. This conflict gives rise to time inconsistent behaviors as well as other violations of rationality. The model presented here nests existing work such as the dual-self \cite{fudenberg2006dual}, and self-control preferences \cite{gul2001temptation}. It can also be thought of as a converged version of a $\mu$Agent model \cite{kurth2009temporal}.

There are three contributions in this paper. First, in many existing models \cite{gul2001temptation,fudenberg2006dual} assume that system 1 is perfectly myopic (aka. cares only about immediate rewards).\footnote{This is not the case for all existing models, for example in \cite{fudenberg2012timing} system 1 has a non-zero discount factor. However, system 1 and system 2 share utility functions and differ only in discount rates. In addition, that work considers applying the model only to simple scenarios rather than arbitrary Markov decision problems.}  Our work extends these models to a system 1 capable of anticipating future consequences of a decision today and adapts standard tools from reinforcement learning to make the model work for arbitrary Markov decision problems. Second, we show how inverse reinforcement learning \cite{ng2000algorithms,ziebart2008maximum}, applied naively, can grossly mislead an analyst trying to learn a 2S agent's goals from their behavior. Third, we adapt a simple IRL algorithm to test for dual-self decision-making or to recover the goals of a dual-self agent from observed behavior.

\subsection{Related Work}
One way to model preference reversals in say that individuals have hyperbolic \cite{ainslie2001breakdown} or quasi-hyperbolic \cite{laibson1997golden,o1999doing} discount rates (i.e. they value $x$ utils in the future as a function such as $\frac{1}{1+kt} x$ instead of $\delta^t x$). In hyperbolic/quasi-hyperbolic models models the effective discounting from waiting from $t=0$ to $t=1$ is larger than the loss from waiting from $t$ to $t+1$. These models display preference reversals, procrastination, and other `irrational' phenomena \cite{o1999doing}. Making dynamic predictions in such models require solving a game between multiple `selves' of an individual that exist at each time period \cite{laibson1997golden,o1999doing,kleinberg2014time,kleinberg2016planning,evans2016learning}.

By contrast, 2S models assume the existence of multiple `selves' (or systems) that exist consistently across periods \cite{kurth2010reinforcement,fudenberg2006dual}. Systems can have different utility functions \cite{hare2009self}, discount the future differently \cite{fudenberg2006dual,kurth2010reinforcement} or have different information sets \cite{kool2018planning}. In each of these models decisions involve some way of combining inputs from these two systems, and they can generate time inconsistent behavior and even discounting patterns that look very similar to those of a hyperbolic discounter \cite{kurth2010reinforcement}.

There are two major advantages of the 2S framework. First, is that it can be applied to decision environments beyond those where decision-makers must trade off rewards now for rewards later. There is strong evidence that system 1 and system 2 appear to differ in their evaluations of many other decisions such as risk/uncertainty \cite{hsu2005neural}, whether to be altruistic \cite{rand2014social,peysakhovich2015habits}, and moral decisions \cite{greene2014moral}. Second, 2S systems give a consistent welfare criterion - they can identify whether an intervention improves or decreases an individual's utility unambiguously (provided we assume system 2 utility is an individual's true utility). Because hyperbolic models assume games between different selves at different time periods asking whether an intervention made an individual better off requires asking which time period's `self' we care about \cite{o1999doing}.

Finally, a growing literature argues that a key distinction between system 1 and system 2 is how they learn \cite{kool2017cost,kool2018planning}. \cite{kool2018planning} argues that system 1 is best modeled as a model-free reinforcement learner while system 2 is more like a model-based planner. While we do not deal with the dynamics of learning in this paper this conception provides a plausible foundation for the assumption that system 1 and system 2 try to maximize different rewards. For example, after trying a few donuts a model free system 1 may learn that donuts are delicious. However, system 1 will not internalize the negative reward of not following a diet because this negative consequence is far into the future. On the other hand, the model-based system 2 can incorporate this reward into it's calculation of an optimal plan. Such dual-system learning patterns have been observed in real world decision-making where information that is personally experienced (e.g. having one's stock portfolio collapse during a recession) has a different effect on decisions than information that is simply learned symbolically \cite{malmendier2011depression}.

\section{Basics of the 2S Model}
We will introduce the 2S model with a simple example and then expand it to Markov decision problems. Consider a decision-maker (DM). The DM is on a diet and is choosing between a delicious but unhealthy donut (d) and a healthy but less flavorful kale smoothie (k) as a snack. 

The decision-maker has $2$ reward functions $r_1$ and $r_2$ which represent systems $1$ and $2$ respectively. System 1 likes sugar whereas system 2's goals include the higher level goal of maintaining the diet. This is formalized as $r_1 (k) = 0$ and $r_1 (d) = 1$ whereas $r_2 (k) = 1$ and $r_2 (d) = -1.$ 

First we consider the DM making a choice for eating the snack right now. Let $a^*$ be the system 1 optimal action (eat the donut). The \textit{cognitive control cost} of deviating to another action $a$ by system $2$ is given by $$CC(a) = \psi (r_1 (a^*) - r_1 (a)).$$ For simplicity let $\psi$ be a linear function. System $2$ trades off its reward and this cognitive control cost, so choices are maximizers of the combined function $$V_{DS} (a) = r_2 (a) -  \psi (r_1 (a^*) - r_1 (a)).$$ Plugging in the rewards above gives $V_{DS} (d) = -1$ and  $V_{DS} (k) = 1 - \psi (1)$

It is easy to see that in this case the DM chooses the kale smoothie iff the cognitive control cost parameter $\psi \leq 2.$ 

Now let us consider the case where the DM is choosing a snack now but will eat it tomorrow ($t=1$). Rewards are received at the time the snack is eaten and system $1$ and $2$ discount future rewards are discounted by rates  $\gamma_1 < \gamma_2.$ System 1 still prefers the donut but now the control cost of deviating is $\psi (\gamma_1 r_1 (a^*) - \gamma_1 r_1 (a))$. Again the choice is the maximizer of the system $2$ utility net of control costs, but now the DM chooses kale when $\psi \leq 2\frac{\gamma_2}{\gamma_1}.$

Thus, there is a range of cognitive control parameters $[2, 2\frac{\gamma_2}{\gamma_1}]$ where agents choose donuts today, but, if they are able to, commit to eating kale tomorrow. 

\section{Planning with Two Systems}
So far we have dealt with a single decision, however, we can consider a 2S DM making a dynamic choice. Suppose that at $t=1$ a donut will be available for lunch and the DM can pay some price at time $t=0$ for kale to be available at $t=1$ as well. Would the DM be willing to pay this price? There are several factors to consider. First, if $\psi > 2$ then even if kale is available the DM would not choose it at $t=1$, thus DM would not be willing to pay at $t=0$. Second, even if the DM would choose kale system 2 must factor into the price the cognitive control costs the DM will have to pay in the future to actually make the choice. Third, if the DM will indeed choose kale at $t=1$ system 1 would prefer that the DM does not make kale available - in other words, in the 2S model there is now a conflict at $t=0$ because of the anticipation of conflict at $t=1.$ 

A Markov decision process (MDP) is a finite set of state $\mathcal{S}$, a set of actions $\mathcal{A}$, a transition function which inputs a state and action and outputs a distribution on the next states $\tau: \mathcal{S} \times \mathcal{A} \to \Delta (\mathcal{S})$. A 2S DM has two reward functions which input a state and action pair and output a distribution on real valued rewards $r_i: \mathcal{S} \times \mathcal{A} \to \Delta (\mathbb{R}).$ Systems discount the future with discount rates $\gamma_1, \gamma_2.$

We now extend the 2S model to MDPs with the formalization. Our DM will be choosing a policy $\pi$ which is a map from states to actions (this can be randomized but in this paper we restrict to deterministic policies). Each system has a value function $V_i (s, \pi)$ which inputs a starting state $s$ and a policy $\pi$ and outputs the expected sum of discounted rewards from behaving according to this policy starting in this state. A related object is each system's $Q$ function $Q_i (s, a, \pi)$ which takes as input a state, action, and policy and gives the expected discounted sum of rewards from taking action $a$ today and following policy $\pi$ starting at the next period. 

Each system has an optimal policy it would prefer, we refer to these as $\pi^*_1$, $\pi^*_2.$ However, behavior will come from system $2$ optimizing its reward function net of cognitive control costs. We call the resulting policy the \textit{compromise policy}. We now turn to understanding what function this policy will actually optimize.

First, we need to ask how to calculate cognitive control costs when we think about policies rather than single actions. As before there is a tradeoff equation for system $2$ given by $$V_{CC} (s, \pi) = V_2 (s, \pi) - CC(s, \pi).$$ We will continue to think about the control costs as the difference between utility gained to system 1 under $\pi$ and $\pi_1^*$ as with the single action case. However, now a choice at time $t$ now affects rewards at $t+k$ and we need to make decisions about how system $1$ perceives the future.

Again, actions are taken at each time step. Let us consider the model where system $1$ suggests an action $\pi^*_1 (s)$. What should the cost be for deviating to a different action?

We now discuss two possible assumptions. We refer to them as the \textit{naive} or \textit{sophisticated} system 1.\footnote{We will use the naive/sophisticated language of the literature on hyperbolic discounting \cite{o1999doing}. In that literature actions of a self in period $t$ depend on expectations of that self about the actions of future selves. Naive agents are those who assume future selves will make the same decisions as the current self, sophisticated agents are those who play a subgame perfect equilibrium (i.e. know that future selves will take actions to maximize their own utility).} 
 
The naive system $1$ assumption is that actions starting tomorrow will follow $\pi^*_1 (s)$ (i.e. system 1 is ignorant of system 2's future plans). A naive system $1$ means that the control cost along a trajectory can be computed via the $Q$ function of system $1$. The per period reward to system $2$ net of control costs can be written as $$r_2 (s, a) - \psi (Q_1 (s, \pi^*_1 (s), \pi^*_1) - Q_1 (s, a, \pi^*_1)).$$  We can thus write the planning problem in standard recursive form 

\begin{align*}
V^{naive}_{CC} (s, \pi) = r_2 (s, \pi(s)) - \\ 
\psi (Q_1 (s, \pi^*_1 (s), \pi^*_1) - Q_1 (s, a, \pi^*_1)) + \\
\gamma_2 V^{naive}_{CC} (s, \pi).
\end{align*} 

Standard RL methods (value iteration, policy iteration) can be used to find the policy which optimizes this function: first we compute $Q_1^*$ using a standard method, then we plug in $Q_1^*$ and again apply policy iteration/value iteration to compute the which optimizes $V^{naive}_{CC}.$ 

The \textit{sophisticated} system $2$ model is more nuanced. Here system $1$ understands that future actions will come from $\pi$ instead of $\pi^*_1$ and so we write the optimizing function net of control costs as $$V^{soph}_{CC} (s, \pi) = V_2 (s, \pi) - \psi (V_1 (s, \pi^*_1) - V_1 (s_, \pi)).$$ The sophisticated formulation is related to the interpretation of 2S model in \cite{thaler1981economic,fudenberg2006dual} where actions are split between a planner and a doer where the only action of the planner is to be able to change the utility function of the doer. In the Markov case this would amount to assuming that the agent starts in state $s$, the planner gets to change the doer's utility function (reward function for each state, action), and the doer chooses the policy consistent with this new utility function.

This no longer has a simple recursive form because $V_1$ and $V_2$ have different discount rates. However, we now show that policy iteration can be used even in this compound problem. First, we substitute the definitions of the value functions into the equations above to re-express $V^{s}_{CC}$ as 
\begin{align*}
 r_2 (s, \pi(s)) + \psi r_1 (s, \pi(s)) + \\ 
  \gamma_2 V_2 (\tau(s, a), \pi) + \psi \gamma_1 V_1 (\tau(s,a), \pi) - \\
\psi V_1 (s, \pi^*_1)
\end{align*}

At each state the optimal policy value for system 1 $V_1 (s, \pi^*_1)$ is a constant. Thus, we can ignore it for the sake of computing the optimal policy starting at that state (though not for computing that policy's actual value). This means that the sophisticated system 1 policy is the one which optimizes the compromise objective $V_2 (s, \pi) + \psi V_1 (s, \pi).$ This also means the cognitive-control based 2S model is a plausible foundation for the $\mu$Agent model \cite{kurth2009temporal}. An adaptation of the value iteration algorithm can be used to construct an optimal policy (Algorithm 1). As with standard value iteration, this algorithm is monotonic (each step has a better policy with respect to the compromise objective) and thus when the MDP is finite it will converge to the optimal policy. 

Note that while standard techniques can be adapted to construct plans for sophisticated system 1 agents their policies no longer have a single associated value function that can be written in standard recursive form. Because system 1 and system 2 have different discount rates the compromise policy may exhibit behaviors like time inconsistency or commitment. Time inconsistency occurs because when an action affects rewards arbitrarily far in the future the discount rate of system $2$ is the only relevant one and thus these decisions will look as if they optimize only for $r_2$ while decisions that affect rewards close in time optimize for a combination of $r_1$ and $r_2.$

\begin{algorithm}
\caption{Value Iteration With Sophisticated System 1}
\begin{algorithmic}

\State Initialize $V_1, V_2$ arbitrarily
\While{Not converged}
\For{$s \in \mathcal{S}$}
\For{$a \in \mathcal{A}$}
\State $Q_1 (s, a) = \mathbb{E} (r_1 \mid s, a) + \gamma_1 V_1 (\tau(s, a))$
\State $Q_2 (s, a) = \mathbb{E} (r_2 \mid s, a) + \gamma_2 V_2 (\tau(s, a))$
\EndFor
\State Let $a^* = \text{argmax}_{a} \psi Q_1 (s,a) + Q_2 (s,a)$
\State Set $V_i (s) = Q_i (s, a^*)$
\EndFor
\EndWhile
\end{algorithmic}
\end{algorithm}

With sophisticated 2S planning in hand, we can move on to examining examples of behavior of 2S agents as well as the differences between sophisticated and naive system 1 agents.

\section{Inverting Dual System Plans}
We now turn to the problem of inferring an agent's reward function from observed behavior. This problem is particularly important if we seek to construct AI that can observe human behavior, infer their desired goal states, and then take actions to help the DM achieve their goals as in, e.g. an artificial assistant.

We consider the standard inverse reinforcement learning (IRL) setup. We have access to a dataset of trajectories (sequences of state-action pairs taken by our agents). We refer to this as $\mathcal{D} = \lbrace \eta_1, \eta_2, \eta_3 \dots \eta_N \rbrace.$ We wish to use $\mathcal{D}$ to infer the underlying reward function(s) of the DM. In addition, we wish to provide a statistical test for the presence of dual-system behavior in a data-set. 

We will do this using maximum likelihood estimation as in prior work \cite{ziebart2008maximum}. We let $\theta$ be the parametrization of the problem (here the reward function or functions). Our goal will be to find the reward parameters $\theta$ to maximize the likelihood of the data. The Markovian property of MDPs means that the likelihood of a trajectory $\eta_i$ can be computed as $$Pr(\eta_i \mid \theta) = \prod_{(s, a) \in \eta_i} Pr(a \mid s, \theta).$$ We will work with the log likelihood instead which will be $$\mathcal{L} (\mathcal{D} \mid \theta) = \sum_{(s,a) \in \mathcal{D}} \text{log} Pr(a \mid s, \theta).$$

With rational agents it is straightforward how to compute $\mathcal{L}$. Given $\theta$ the agent has an optimal policy which has an associated $Q$ function. We refer to this as $Q^* (s, a \mid \theta)$. We assume that choices at each state are made according to a softmax of this $Q^*$ function $$Pr(a \mid s, \theta) = \dfrac{\text{exp} (\frac{1}{\beta} Q^* (a, s \mid \theta))}{\sum_{a'} \text{exp} (\frac{1}{\beta} Q^* (a', s \mid \theta))}.$$ Thus the log likelihood is well defined. 

We now turn to, given a guess $\theta$ calculating the likelihood of observed trajectories for 2S agents. We begin with the naive system $1$ agents. This is the easier of the two cases. Recall that here we can think of this agent as a rational agent which optimizes a new reward function which takes the discounted (with system 2 discount rate) payoffs of the form $$r^{naive} (s, a) = r_2(s, a) - \psi (V^*_1 (s) - Q^*_1 (s, a)).$$ Thus we can readily compute a $Q^{naive*}_{CC}$ function if we can compute the optimal policy. This gives us a well defined likelihood for any trajectory.

For sophisticated system 1 agents the problem is a little bit more complicated. Recall that a sophisticated system 1 agent's final policy also optimizes the compromise value function $V_2 (s, \pi) + \psi V_1 (s, \pi).$ We can can write the associated $Q$ function as

\begin{align*}
Q^{soph}_{CC} (s, a, \pi) = r_2 (s, a) + r_1 (s, a) + \\
\gamma_2 V_2 (\tau(s,a), \pi) + \gamma_1 V_1 (\tau(s,a), \pi).
\end{align*} 

Given the policy $\pi^*_{soph}$ we can compute this $Q$ function and plug it into the softmax equation above. Now we have a well defined likelihood given the parameters $\theta.$ 

\section{Experiment: Donut Kale Grid World}
\begin{figure}[ht!]
\begin{center}
\includegraphics[scale=.32]{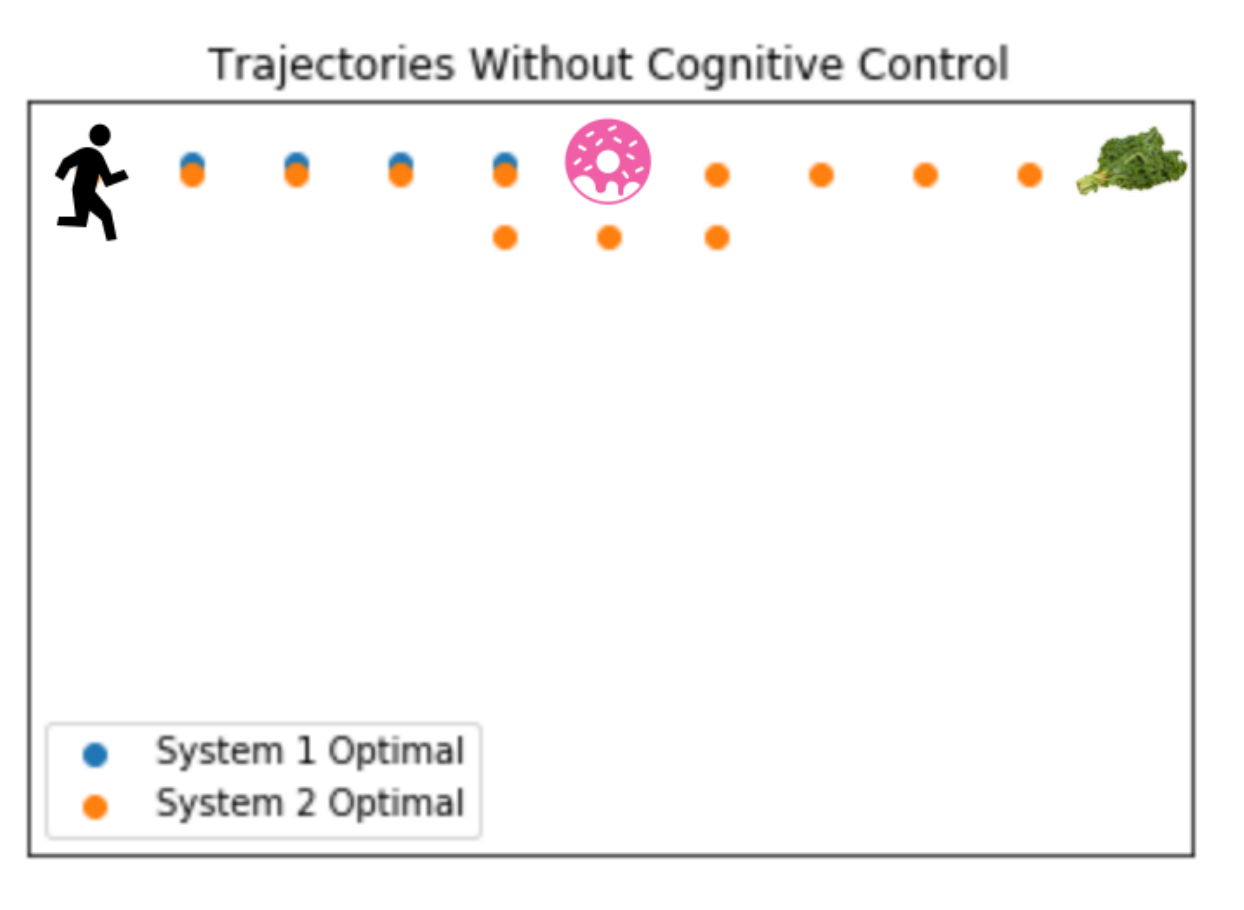} 
\includegraphics[scale=.32]{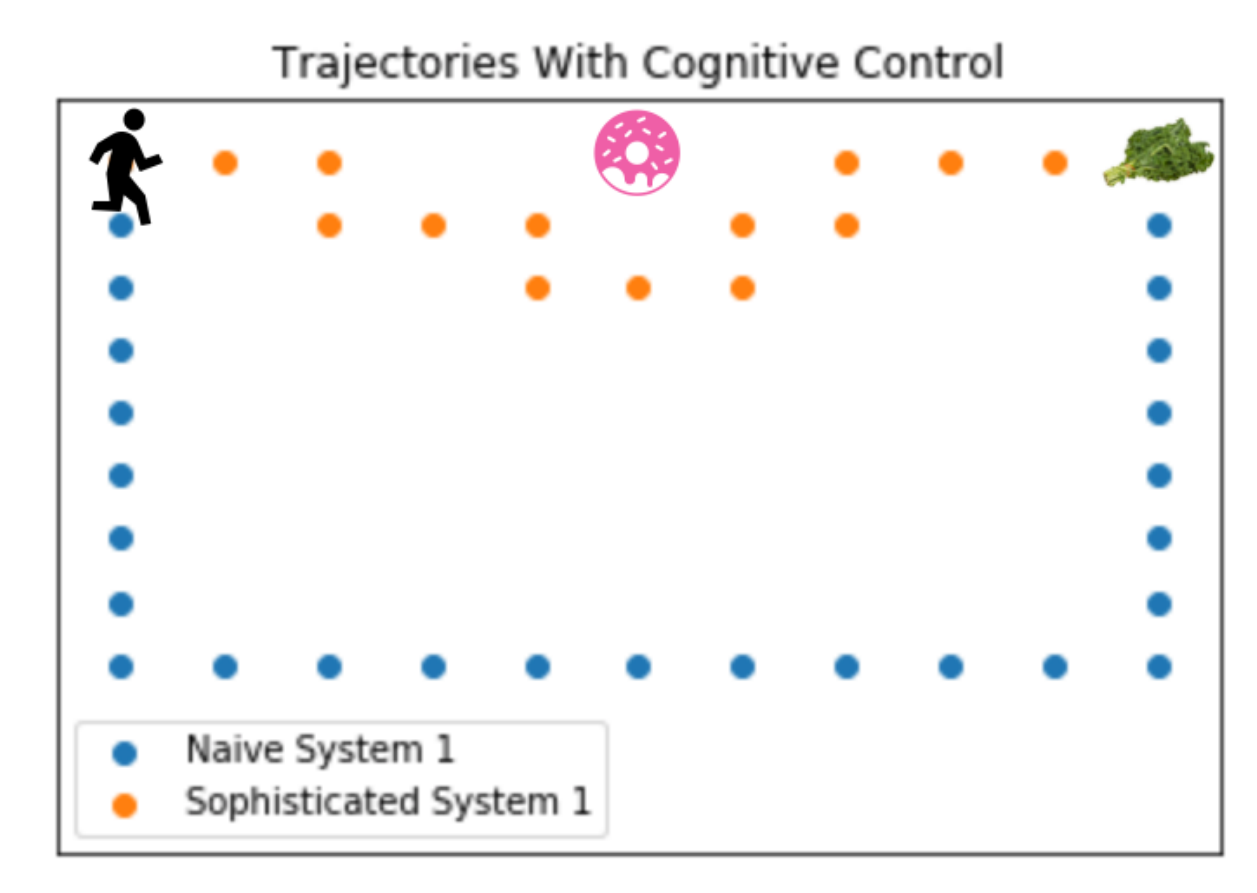} 
 \includegraphics[scale=.32]{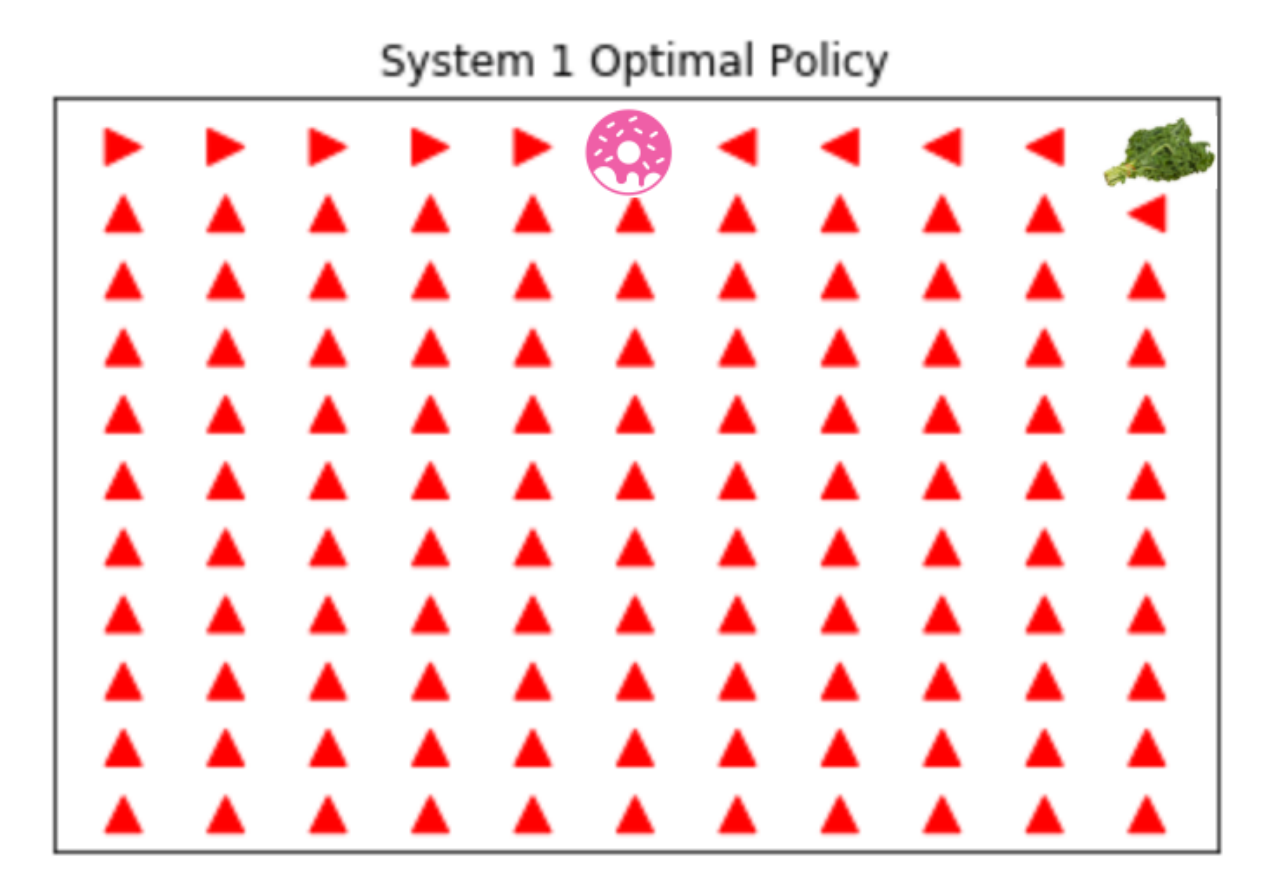} \includegraphics[scale=.32]{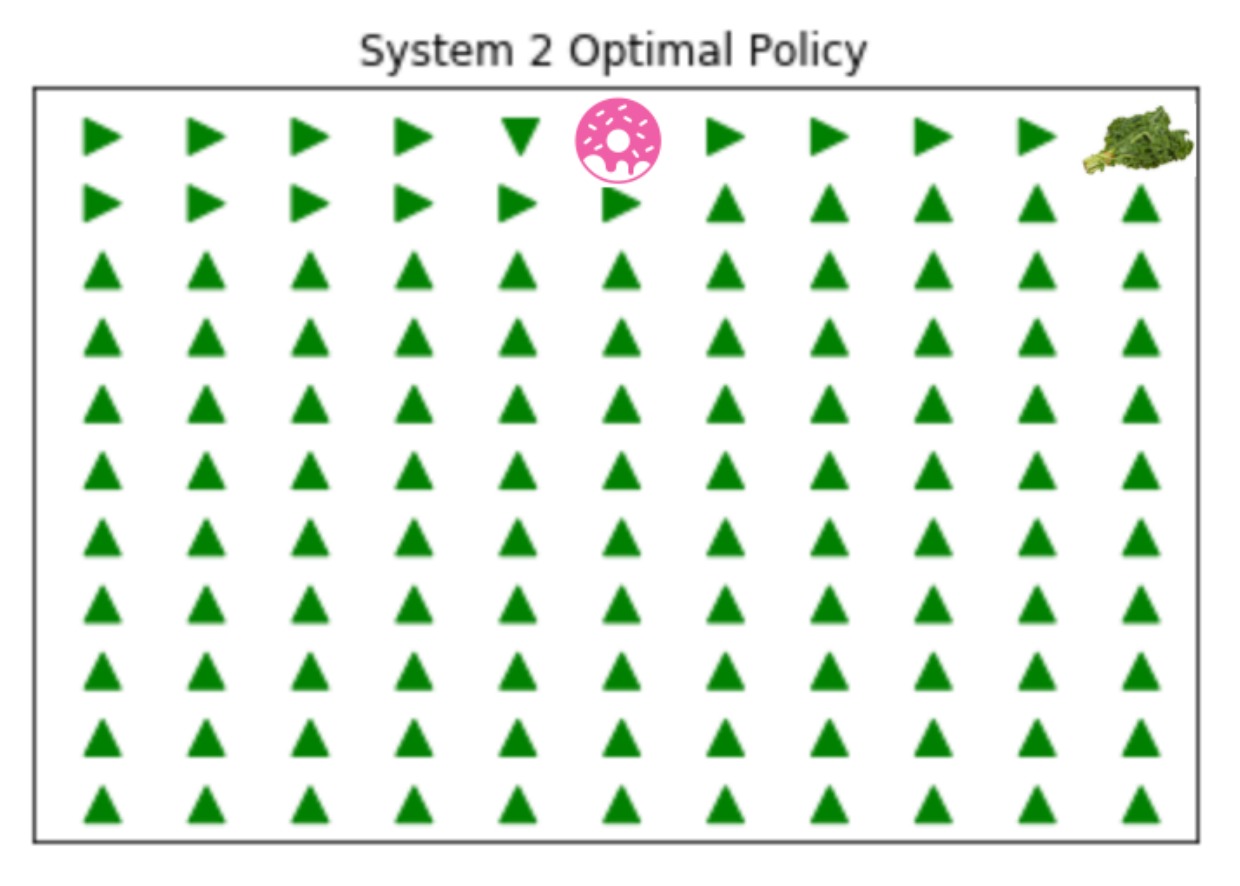}
 \includegraphics[scale=.32]{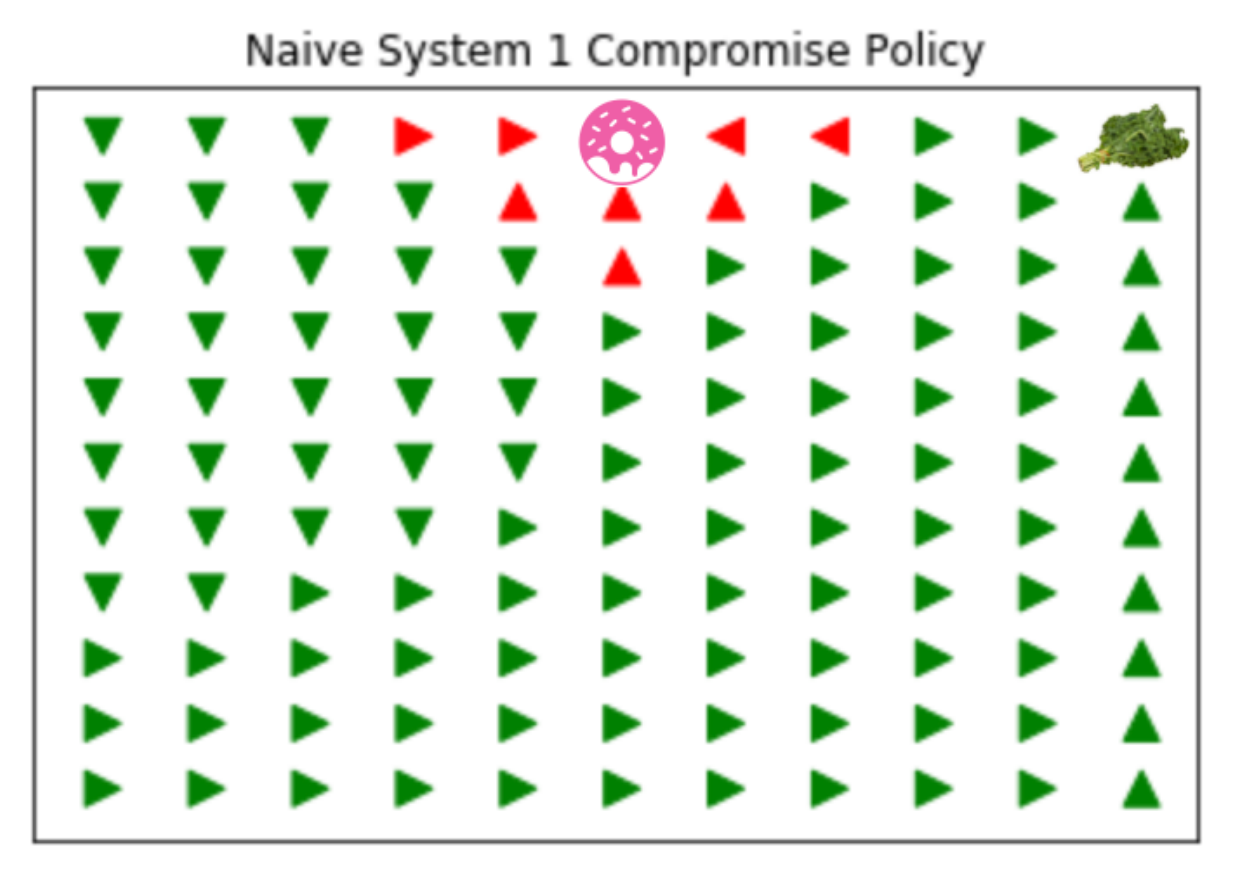} \includegraphics[scale=.32]{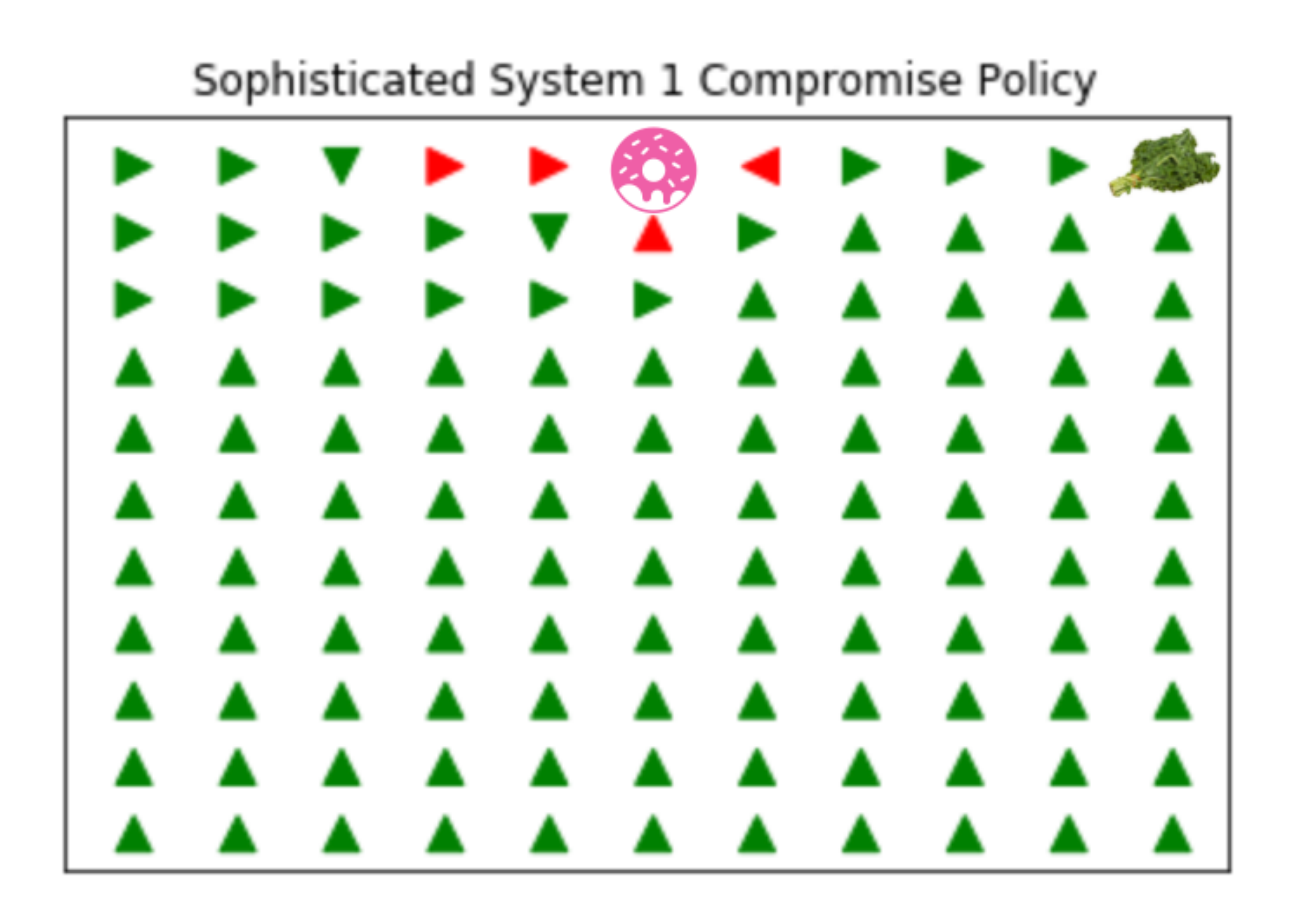} \includegraphics[scale=.4]{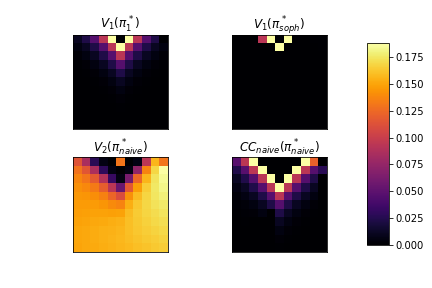} 

\end{center}
\caption{Top row shows the trajectories of system 1 and system 2 starting from the corner as well as the full preferred policies of both systems. Each arrow indicates the direction the agent will move at that state. States are colored red if starting at that state leads to the donut and green if it leads to the kale. Bottom row shows the compromise policies under the 2S model with naive/sophisticated system 1. The compromise policies are not pure mixtures of the optimal policies and display behavior such as precommitment (walking far away from the donut to make it harder to reach in the future). Last subpanel shows functions for each of the systems given compromise policies as well as the map of self control costs incurred by the DM in each state under the naive assumption}
\label{grid_world_figure}
\end{figure}

We consider a Markov grid world. An agent begins in a location and can move in any of $4$ cardinal directions. There are two items placed on the grid, a kale smoothie and a donut (see Figure \ref{grid_world_figure}). The donut and kale are terminal states of the game. The donut and kale have rewards for system $1$ and system $2$ as in the example above. We set $\gamma_2 = .99, \gamma_1 = .6, \psi = 5.$

Figure \ref{grid_world_figure} shows the world as well as the trajectories starting from the corner of the board. The ideal policies of system 1 (grab the donut) and system 2 (walk to the kale using the shortest possible path). However, the compromise policies look quite different from either optimal policy. Indeed, both of them take a path around the donut (but not the shortest possible path). We see the naive and sophisticated system 1 policies differ quite a lot with the compromise policy with the naive system 1 demonstrating giving the donut a very wide berth. 

The difference in final behavior under the naive/sophisticated assumptions comes down to the following logic. Consider a simple problem where the DM chooses between $2$ actions Stop and Go at $t=0$. If the agent Stops the game ends, otherwise it continues to $t=1$ where the agent also chooses between Stop and Go. If the agent chooses Stop, the game ends, if the agent chooses Go, he gets a donut. System 1's optimal policy is to Go at both stages and eat the donut. Suppose the cognitive control parameters are such that agents Stop in period $t=1$ (with a control cost). A sophisticated system 1 then knows at $t=0$ that the agent will stop and therefore there is no control cost for Stop in period $0$. However, a naive system 1 always assumes that the agent will Go in the future and therefore there will be a control cost at period 1. This reasoning is precisely why the DM in the grid world gives the donut a wide berth under the naive system 1 assumption but walks exactly one step away under the sophisticated assumption.

Whether the naive or sophisticated model of cognitive control is more reflective of real decision-making is an open empirical question (though cognitively it seems like the naive assumption is more plausible), however, because the models give such precise predictions about the main differences it is an empirically testable one and a potentially fruitful direction for future research.
 
\section{Experiment: IRL in the Donut Kale Grid World}
We now apply the IRL methods above to the Donut Kale grid world problem. We keep the same true underlying problem parameters and use the true value functions to construct a dataset of stochastic trajectories starting from any possible (non-terminal) state. We construct datasets of stochastic trajectories for both kinds of dual system agents. We sample $50$ stochastic trajectories using softmax choice of actions at each time point with $\beta=.01$ starting from any possible non-terminal state. We give models access to the true discount rates. We use a large number of trajectories as well as the true discount rates because the purpose of this experiment is to show that rational IRL draws problematic inferences even when it has access to large amounts of data and true discount rates. We refer to these data sets as $\mathcal{D}_{naive}$ and $\mathcal{D}_{soph}.$ 

For each of the different assumptions on the true planning function (rational, sophisticated, naive) optimization of the IRL likelihood function is relatively slow (because each iteration requires the computation of an optimal policy). In these experiments we found that the IRL objective functions are very difficult to optimize, methods that improved the likelihood function only locally ended up finding very poor minima and we ended up using differential evolution to optimize the likelihood \cite{storn1997differential}. In standard rational IRL there exist assumptions which allow for closed form solutions to gradients at each optimization step \cite{ziebart2008maximum}. Unfortunately these assumptions do not hold in the 2S case, and so an open area for future research is the efficient computation of solutions to dual-system IRL models.

We the rational, sophisticated, and naive IRL models using maximum likelihood estimation on the generated datasets. We find that a rational IRL model recovers that both 2S agents like donuts (this is because if the DM starts close to the donut then they eat it, thus the only way to rationalize this is to have positive reward from the donut). Thus, using rational IRL would lead a designer to make wrong decisions if they could make interventions (for example, if they are asked whether the DM would prefer a donut or nothing). 

Second, we find that correctly specified models learn the underlying preferences quite well. Both the naive and sophisticated IRL learn that only system 1 prefers donuts. Things are more mixed when we examine misspecified 2S IRL models. The naive IRL model applied to the sophisticated data yields wrong inferences just like the rational model, however the sophisticated model applied to the naive data gets the order of preferences correct.

\begin{figure}[h]
\begin{center}
\includegraphics[scale=.4]{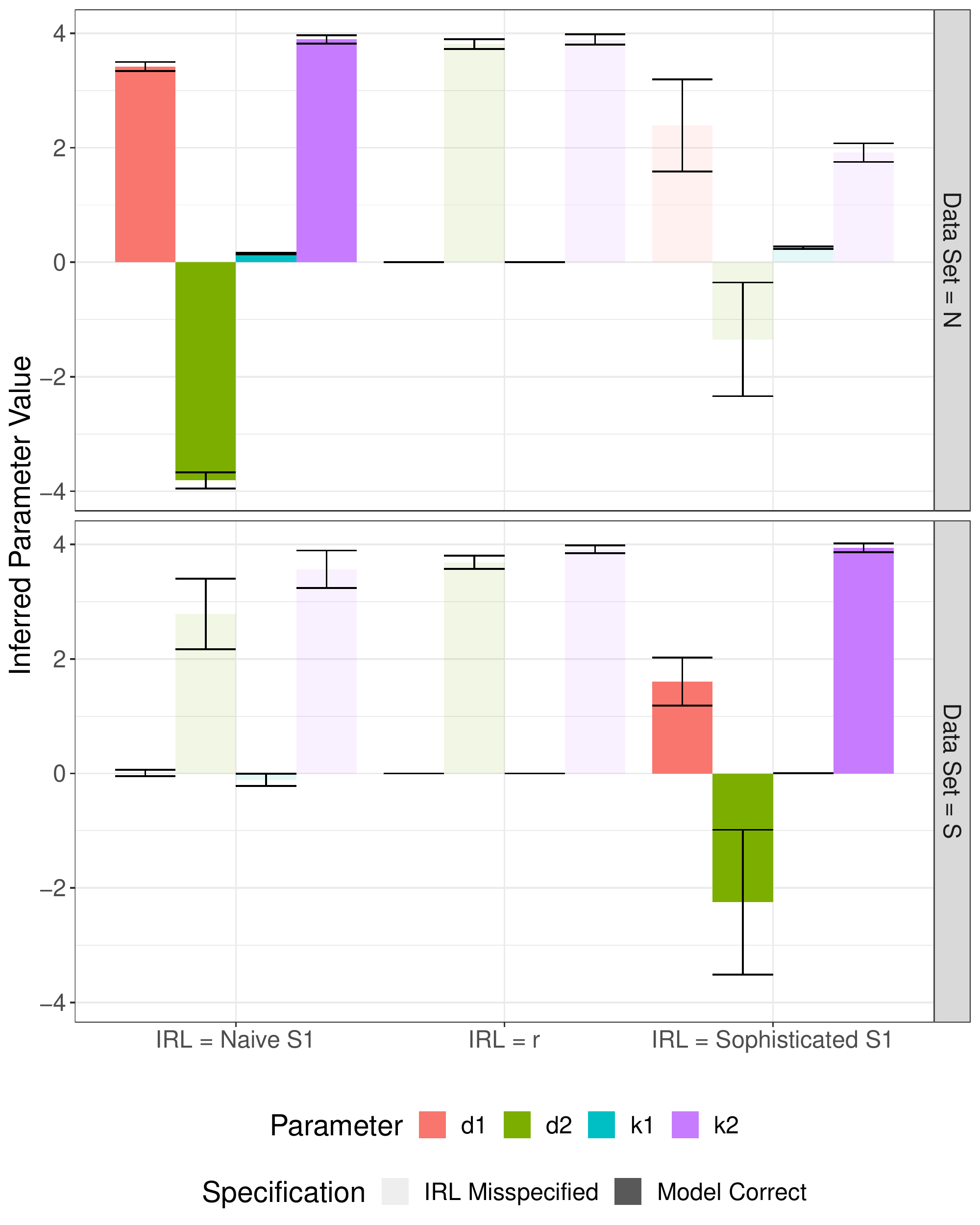}

\end{center}
\caption{Rational IRL makes incorrect inferences about the decision-maker's preferences (d1, d2, k1, k2 refer to inferred rewards of donuts, kale for system 1 and system 2 respectively). However, 2S IRL models with correct specification learn that only system 1 likes donuts (i.e. $d1 > 0, d2 < 0$). Misspecification is more of a mixed bag with the sophisticated model learning the correct preferences when applied to the naive dataset but the naive model inferring the wrong answer. Lines reflects averages over 16 replicates, error bars reflect standard deviations.}
\label{irl}
\end{figure}

\section{Conclusion}
This work has focused on extending planning and inverse planning models into the domain of non-rational agents. In particular we have focused on the dual system framework that has proven to be successful across the behavioral sciences. We have shown that standard planning and inverse planning algorithms can be adapted to 2S agents. Importantly we have shown that incorrectly assuming that a dual-system agent is a rational agent when trying to infer goals from behavior can lead to interventions that actually decrease the overall welfare of the agent.

We have assumed that both system 1 and system 2 have fixed reward functions and use model-based methods for computing optimal polices. In reality, both systems are constantly learning about the real world. Existing work \cite{kool2017cost,kool2018planning} argues that these systems learn and plan differently with system 1 being more model-free and system 2 being more model-based. Extending the 2S model presented here to include both general reinforcement learning and known regularities in human learning \cite{erev1998predicting,hertwig2004decisions,fudenberg2016recency} is an important future direction.

We have focused on the domain of time inconsistency but system 1/system 2 conflicts occur in many other domains. Extending the model here to problems in moral decision-making, decision making under risk and uncertainty, and cooperation sees like a fruitful direction for future research. In particular, recent work in behavioral science argues that cooperation is an interplay between a system 1 which learns `social heuristics' and a reward maximizing system 2 \cite{rand2012spontaneous,rand2014social}. However, formal models of the social heuristics hypothesis (SHH) have been restricted to simple matrix games \cite{bear2016intuition}. The 2S model presented here is a potential way to expand the SHH to more complex environments such as Markov social dilemmas \cite{leibo2017multi,lerer2017maintaining,peysakhovich2017consequentialist}.

Finally, rational actor models are used explicitly or implicitly across many applications (a large literature in recommender systems can be framed as a form of learning latent preferences). Understanding where such models are appropriate and where the assumptions are so badly broken that they lead to false conclusions is an important topic for discussion, research, and debate. 

\bibliography{20190304dualsystem_AIESFinal.bbl}

\bibliographystyle{ACM-Reference-Format}

\end{document}